%% file: main.tex
\documentclass{article}
\usepackage{iclr2026_conference}
\usepackage{times}

\input{math_commands.tex}


\usepackage{hyperref}
\usepackage{url}
\usepackage{graphicx}
\usepackage{booktabs}
\usepackage{siunitx}
\usepackage{svg} 
\usepackage{lipsum}
\usepackage{amsthm}
\usepackage{mathtools}
\usepackage{amssymb}
\usepackage{subcaption}
\usepackage{comment}
\usepackage{enumitem}
\usepackage[normalem]{ulem}

\DeclarePairedDelimiter{\norm}{\lVert}{\rVert}

\graphicspath{{imgs/}}

\title{Accelerated Gradient Descent for Faster Convergence with Minimal Overhead}
\author{Manuel Graca\thanks{\texttt{manuel.s.graca@tecnico.ulisboa.pt}}\\
IST, ULisboa\\
Lisbon, Portugal\\
  \And
L. Miguel Silveira\\
IST, ULisboa\\
Lisbon, Portugal\\
\And
Arlindo Oliveira\\
IST, ULisboa\\
Lisbon, Portugal\\
\And
Frank Liu\textsuperscript{\dag} \\
\textsuperscript{\dag}Old Dominion University \\
Norfolk, VA USA\\
}

\begin{document}
\maketitle

\begin{abstract}
\input{sections/00_abstract}
\end{abstract}

\input{sections/01_introduction}
\input{sections/02_method}

\input{sections/03_tool}

\input{sections/04_results}

\input{sections/05_conclusion}

\newpage
\bibliography{bib/iclr2026_conference}
\bibliographystyle{iclr2026_conference}
\newpage

\appendix
\renewcommand{\thesection}{S}
\input{sections/A1_appendix}

\end{document}

%% file: math_commands.tex
\usepackage{amsmath,amsfonts,bm}
\usepackage{algorithm}
\usepackage{algpseudocode}
\algrenewcommand\algorithmicrequire{\textbf{Input:}}
\algrenewcommand\algorithmicensure{\textbf{Output:}}
\algrenewcomment[1]{\hfill\(\triangleright\) #1}

\usepackage{subcaption}
\usepackage[table]{xcolor}
\definecolor{curvesel}{gray}{0.92}

\usepackage{adjustbox}

\newif\ifshowrevisions
\showrevisionstrue      
\showrevisionsfalse   

\ifshowrevisions
  \newcommand{\rev}[1]{{\color{blue}#1}}          
  \newcommand{\rem}[1]{{\color{blue}\sout{#1}}}   
\else
  \newcommand{\rev}[1]{#1}  
  \newcommand{\rem}[1]{}    
\fi

\newcommand{\opt}{CT-AGD}

\def\Secref#1{Section~\ref{#1}}

\def\eqref#1{equation~\ref{#1}}
\def\Eqref#1{Eqn.~\ref{#1}}

\def\1{\bm{1}}

\DeclareMathAlphabet{\mathsfit}{\encodingdefault}{\sfdefault}{m}{sl}
\SetMathAlphabet{\mathsfit}{bold}{\encodingdefault}{\sfdefault}{bx}{n}

\newcommand{\E}{\mathbb{E}}

\newcommand{\R}{\mathbb{R}}

\DeclareMathOperator*{\argmin}{arg\,min}

%% file: sections/00_abstract.tex
In this paper, we present \opt\ (Curvature-Tuned Accelerated Gradient Descent), an optimization method for non-convex optimization problems in deep learning training tasks. \opt\ is a general boosting procedure that accelerates first-order methods by explicitly capturing the local curvature using finite-difference quotients, and the development of heuristics aimed at mitigating noise and bias introduced by stochastic mini-batch training. \opt\ has a comparable storage and computational overhead as adaptive gradient methods such as Adam. Our extensive experiments demonstrate that \opt\ achieves the same level of accuracy as the baseline first-order methods, yet reduces the required training epochs by $33\%$ on average.

%% file: sections/01_introduction.tex
\section{Introduction}
\label{sec:intro}

We investigate the optimization methods in the training of deep neural networks:
\begin{equation}
\label{eqn:nn_training}
\theta^\star
    = \arg\min_{\theta \in \mathbb{R}^d}
      \mathcal{L}(\theta)
    = \arg\min_{\theta \in \mathbb{R}^d}
      \mathbb{E}_{(x,y)\sim \mathcal{P}}
      \bigl[\,\ell(f_\theta(x), y)\,\bigr]
\end{equation}
where $f_\theta : \mathcal{X} \rightarrow \mathcal{Y}$ represent the neural network, $\theta \in \mathbb{R}^d$ are trainable parameters, $\ell (\cdot)$ is the point-wise loss function, and $\mathcal{P}$ is the data distribution. The core of training tasks is essentially the optimization of a large-scale non-convex loss function, requiring substantial computational resources and energy~\citep{Strubell2019Energy,Schwartz2020GreenAI, Patterson2022Carbon}. Despite significant progress in accelerator hardware~\citep{Sze2020Efficient, Jouppi2017TPU, HennessyPatterson2019DSA}, the optimizers remain critical to reduce end-to-end training time while ensuring stable convergence~\citep{BottouCurtisNocedal2018}.

Although higher-order, particularly second-order, optimization methods can achieve faster convergence under smoothness and convexity assumptions~\citep{Nesterov2004,BoydVandenberghe2004,Bubeck2015}, their theoretical advantages are often diminished in deep learning training due to the non-convexity of the loss function landscape, as well as stochastic noises from mini-batch sampling and data augmentation drift~\citep{ADAHessian2021,pmlr-v80-bollapragada18a}. First-order methods (SGD and its variants) remain popular in deep learning training due to their robustness and relatively low computational cost~\citep{Goyal2017Imagenet1Hr,Smith2018DontDecayBatchSize}.

Adaptive gradient methods using exponential moving averages have
shown great success in addressing some key shortcomings of the first-order optimization methods,
particularly poor convergence due to the noisy gradient from mini-batching sampling and the
sensitivity of learning rates to different scales of widely different curvature. Among them,
ADAM~\citep{Kingma2015Adam}, RMSProp~\citep{Tieleman2012RMSProp} and
ADADelta~\citep{Zeiler2012Adadelta} have been deployed in the successful training of many
applications. However, as pointed out in~\citet{Reddi2018AMSGrad, Zaheer2018AdamNC}, ADAM-family of adaptive
gradient methods may fail to converge to the optimal solution in some relatively simple convex
optimization problems when a constant mini-batch size is used. Improvements were also proposed
in~\citet{Zaheer2018AdamNC} to address the convergence issues with better adaptivity of the
learning rate. Implicitly, the adaptive gradient methods can be considered as an approximation of
the diagonal entries of the full Hessian matrix (or {\it local curvature}), and use the
information to provide better learning rates for different dimensions, versus a global learning
rate in the first-order methods. 

In this paper, we take a more direct approach by calculating approximations to the diagonal entries of the
second-order derivatives (or \emph{quotients}, or \emph{second differences}) during the training
steps. Our proposed method, \emph{Curvature-Tuned Accelerated Gradient Descent} or \opt,
accumulates the second differences that are lightweight approximations of the local
curvature. Within each epoch, \opt\ behaves the same as a standard first-order optimizer. In
parallel, \opt\ accumulates per-coordinate quotients, which serve as approximations of local
curvatures. Clamp functions are applied to guard the quotients into
positive intervals, which suppress oversized steps, and to ensure optimization is
stable. At the end of each epoch, \opt\ performs one additional first-order update, using the
quotient accumulation, which can be viewed as the approximation of the diagonal of the
Hessian, and provides a direct measure of the local curvature. These direct estimates of the local
curvature are also used in the subsequent epoch for faster convergence.

\textbf{Contributions} The main contributions of this paper are as follows:

\begin{enumerate}[noitemsep, topsep=0pt, leftmargin=*]
    \item We introduce a simple, memory- and compute-efficient optimizer, \textbf{\opt}, that can be deployed with existing first-order methods (e.g., SGD, ADAM, etc) with a once-per-epoch \rem{once-per-epoch diagonal quotient preconditioner} \rev{curvature estimation}. By deploying \opt\, the optimization method can capture the local curvature while preserving first-order scalability.      
    \item We conduct extensive numerical experiments on the effectiveness and efficiency of \opt. We also provide an intuitive visualization and benchmarking testbed on DL-like objectives. We hope this testbed can be valuable in supporting transparent and reproducible evaluation of future research activities in this direction. 

\end{enumerate}

\subsection*{Related Work}
\label{subsec:related}

\textbf{First-order methods.}
Stochastic gradient descent (SGD) remains the cornerstone of large-scale NN training because of its low per-iteration cost and strong anytime performance. Practical accelerations include momentum \citep{Polyak1964Momentum} and Nesterov’s acceleration \citep{Nesterov2004}, as well as adaptive methods that maintain running statistics of gradients to adjust coordinate-wise step sizes (without modelling curvature explicitly). Representative examples are AdaGrad \citep{Duchi2011AdaGrad}, RMSProp \citep{Tieleman2012RMSProp}, AdaDelta~\citep{Zeiler2012Adadelta}, AdamW~\citep{LoshchilovHutter2019AdamW}, NAdam~\citep{Dozat2016NAdam}, Yogi~\citep{yogi2018} and Adam \citep{Kingma2015Adam}. Empirically, such methods are competitive across many workloads, but can be sensitive to hyperparameters, and they may converge to an approximate second-order critical point~\citep{Wilson2017Marginal,Ruder2016Overview,Keskar2017LargeBatch,pmlr-v84-reddi18a,levy2016power,pmlr-v70-jin17a,pmlr-v40-Ge15}.
Nonetheless, rigorous nonconvex analyses have been done recently for many of these methods~\citep{Reddi2018AMSGrad,WardWuBottou2019Adagrad,yogi2018}. Our work follows this line but uses a \emph{quotient-based} diagonal scaling tied explicitly to curvature proxies gathered during the preceding epoch.

\textbf{Second-order, natural-gradient, and quasi-Newton methods.}
Quasi-Newton methods approximate inverse curvature from observed changes during optimisation. The canonical example is limited-memory BFGS (L\mbox{-}BFGS), which stores a short history of curvature pairs $(s_i,y_i)$ to build a low-rank inverse–Hessian and applies it with the two-loop recursion without forming matrices \citep{LiuNocedal1989,Schraudolph2007SQN,WangMaGoldfarb2017}. In deep learning, high dimensionality and mini-batch noise motivate stochastic variants that add damping to control indefiniteness, partition parameters into blocks to reduce memory, and refresh curvature only periodically. These variants can also use Hessian–vector products to apply curvature implicitly, preserving tractability while retaining curvature-informed directions \citep{Pearlmutter1994,Byrd2016StochasticQN,Goldfarb2020Practical,Anil2020Shampoo,Gupta2018Shampoo, ADAHessian2021}.

A complementary line of work replaces the Hessian with statistical surrogates. Natural gradient uses the Fisher information as a metric and preconditions gradients by an estimate of $F^{-1}$, which improves conditioning and provides parameterisation invariance \citep{Amari1998}. Hessian-free and generalised Gauss--Newton methods construct a positive semidefinite curvature matrix such as the GGN or Fisher, and compute search directions with conjugate gradient using Hessian–vector products \citep{Martens2010HF,Schraudolph2002GGN}. Kronecker-factored approximations such as K-FAC approximate layer-wise Fisher blocks with Kronecker products of small factors, yielding a tractable block-diagonal preconditioner that lowers memory and compute cost and scales to large models \citep{MartensGrosse2015KFAC,GrosseMartens2016}.

%% file: sections/02_method.tex
\section{Loss Optimization in Neural Networks}
\label{sec:loss-optimization}

\subsection*{Problem Statement}
\label{subsec:problem}

Consider the training task of a deep learning network as described in Eqn.~(\ref{eqn:nn_training}).
Let the training set be $\{(x_i,y_i)\}_{i=1}^I$. 
Consider a depth-$L$ feedforward network with parameters
\begin{equation}
\theta = \{W_\ell, b_\ell\}_{\ell=1}^L,
\end{equation}
where each layer $\ell$ has a weight matrix $W_\ell$ and a bias vector $b_\ell$.
In the forward pass step, given an input $x$, the activations are defined recursively:
\begin{equation}
a_0 = x, \qquad 
h_\ell = W_\ell a_{\ell-1} + b_\ell, \qquad 
a_\ell = \phi_\ell(h_\ell), \quad \ell = 1,\dots,L.
\end{equation}
Here $h_\ell$ are the \emph{pre-activations} (linear combinations of inputs plus bias), 
and $\phi_\ell(\cdot)$ is a chosen nonlinear activation function.
For a per-example loss $\ell(z,y)$ evaluated at the output $z=a_L$, 
backpropagation computes derivatives backward as follows:
\begin{equation}
\mathcal{D}a_L \;\; \leftarrow \;\; \left.\frac{\partial \,\ell(z,y)}{\partial z}\right|_{z=a_L},
\end{equation}
and for $\ell = L,\dots,1$:
\begin{equation}
g_\ell = \mathcal{D}a_\ell \odot \phi_\ell'(h_\ell), \qquad
\nabla W_\ell = g_\ell a_{\ell-1}^\top, \qquad
\nabla b_\ell = g_\ell, \qquad
\mathcal{D}a_{\ell-1} = W_\ell^\top g_\ell.
\end{equation}

\subsubsection*{First- vs Second-Order Optimization Methods (with Adam/Yogi)}
First-order algorithms such as SGD, and the variants with EMA enhancements such as RMSProp, and Adam update parameters using stochastic gradients,
\begin{equation}
\theta_{t+1} = \theta_t - \eta \nabla L(\theta_t),
\end{equation}
with adaptivity and momentum altering the effective step sizes and directions. These methods are memory efficient and map well to accelerators and mini-batch pipelines, which is why they dominate large-scale training. Their main limitation is sensitivity to ill-conditioning and plateaus in the loss landscape~\citep{Ruder2016Overview,ADAHessian2021}. By contrast, classical second-order methods like Newton and Levenberg--Marquardt incorporate curvature via (approximations to) the Hessian and can converge rapidly near minima~\citep{HaganMenhaj1994}. In deep learning, however, forming or inverting dense curvature is prohibitive, and the stochasticity of mini-batches undermines accurate curvature estimation, which diminishes their practical effectiveness at scale~\citep{AmpazisPerantonis2002}.
Structured approximations (e.g., block diagonal and Kronecker factorizations\rev{~\citep{MartensGrosse2015KFAC}}) and stochastic quasi-Newton variants narrow the gap but still face stability and overhead trade-offs in realistic DNN 
training tasks.

\paragraph{L\mbox{-}BFGS in Brief}
Limited-memory BFGS (L\mbox{-}BFGS) maintains a short history of curvature pairs $(s_\ell, y_\ell)$ with $s_\ell=\theta_{\ell+1}-\theta_\ell$ and $y_\ell=\nabla L(\theta_{\ell+1})-\nabla L(\theta_\ell)$, defining an implicit inverse-Hessian $H_t^{-1}$. The two-loop recursion computes the search direction,
\begin{equation}
p_t \;=\; -\,H_t^{-1}\,\nabla L(\theta_t), \qquad
\theta_{t+1} \;=\; \theta_t \;+\; \alpha_t\, p_t \;=\; \theta_t \;-\; \alpha_t\, H_t^{-1}\,\nabla L(\theta_t).
\end{equation}
The two-loop recursion applies $H_t^{-1}$ to a vector in $O(r\,d)$ time using $r$ stored pairs and $d$ parameters (with $r$ typically $5$–$20$ in deep nets). The step length $\alpha_t$ is then chosen separately, usually by a line-search which requires $K-1$ extra forward–backward passes.

\paragraph{Adam/Yogi in Brief}
Adam and Yogi maintain exponential moving averages of the gradient and its squared magnitude, \rev{with $g_t=\nabla L(\theta_t)$,}\rem{yielding a diagonal preconditioner. With $g_t=\nabla L(\theta_t)$},
\begin{eqnarray}
m_t=\beta_1 m_{t-1}+(1-\beta_1)g_t, & \\ \nonumber
& \hspace*{-1.75in} \text{Adam: } v_t=\beta_2 v_{t-1}+(1-\beta_2)g_t^2,\quad
\text{Yogi: } v_t=v_{t-1}-\operatorname{sign}(v_{t-1}-g_t^2)(1-\beta_2)g_t^2.
\end{eqnarray}

After bias correction $(\hat m_t,\hat v_t)$, the \rem{preconditioned} search direction and update are
\begin{equation}
\label{eq:adam_update}
p_t \;=\; \frac{\hat m_t}{\sqrt{\hat v_t}+\varepsilon}, \qquad
\theta_{t+1} \;=\; \theta_t \;-\; \eta\, p_t .
\end{equation}
Per-step memory and compute are $O(d)$. The step length $\eta$ is set externally (often with a schedule), rather than via a line search. AdamW decouples weight decay from the adaptive update; Yogi’s variance rule caps unwarranted growth of $v_t$ under noisy gradients.

\subsection{The \opt\ Method}\label{subsec:opt-formal}

The rationale of \opt\ is to provide more accurate estimates of local curvatures to the optimizers. It can be interpreted as a local-curvature-aware booster of first-order optimizers. Within each mini-batch, it estimates the diagonal of the Hessian, $\widehat H$, using finite-difference quotients. At the end of each epoch, \rev{it uses this information to perform a more informed update}\rem{computes a \emph{preconditioned} update}.

\noindent\textbf{Conventions.}
Unless stated otherwise, absolute values, inequalities, divisions, and the clamp $\Pi_{[\scriptstyle \lambda_{\mathrm{min}},\lambda_{\scriptstyle \mathrm{max}}]}(\cdot)$ act element-wise; $\oslash$ denotes the element-wise division and $\odot$ denotes the Hadamard product. $K$ denotes the number of epochs with $k$ as the epoch index. $T$ denotes the number of mini-batches in each epoch, with $t$ as the index. Within each mini-batch, the training samples are drawn randomly without replacement with the batch size of $B$. 

\textbf{Within-Epoch First-Order Steps}
In \opt\ , the optimization within each epoch is almost identical to the classic first-order
methods. Therefore, different first-order methods (e.g., SGD and Adam) can be used as the backbone. To simplify annotation, we use vanilla SGD as the example.
Starting from $x_{k,0}$, the inner iterations for $t=0,\dots,T-1$ are done as follows:
\begin{equation}
\label{eqn:dk_schedule}
\theta_{k,t+1}\;=\;\theta_{k,t}-\mu_{k,t}\, g_{k,t},
\qquad
\mu_{k,t}\;=\;\frac{\eta_1}{\gamma_{k,t}},
\end{equation}
with a scalar, iteration-varying \emph{curvature-aware divisor}
\begin{equation}
\gamma_{k,t} = \gamma_k-\bigl(\gamma_k-1\bigr)\frac{t}{T},\qquad t=0,\ldots,T.
\end{equation}
With the divisor at the endpoints: \(\gamma_{k,0}=\gamma_k\), \(\gamma_{k,T}=1\). (See \ref{subsec:curvature_gamma} for a plot of \(\gamma_{k,t}\) over an epoch.)

$\gamma_k$ comes from the curvature estimation of the previous epoch. However, with every step, it loses fidelity as the inner steps move further away from the point of the curvature estimation, hence throughout the epoch: 
\begin{equation}
\begin{cases}
\gamma_k>1: & \mu_{k,t}\text{ starts low and increases linearly to }\eta_1,\\
0<\gamma_k<1: & \mu_{k,t}\text{ starts high and decreases linearly to }\eta_1,\\
\gamma_k=1: & \mu_{k,t}\equiv\eta_1.
\end{cases}
\end{equation}
\noindent\emph{Interpretation:} $\gamma_k>1$ indicates high curvature; $0<\gamma_k<1$ indicates low curvature.

\textbf{Direct Accumulation of Diagonals of Hessian}
Define first differences for $t=1,\dots,T-1$,
\begin{equation}
\Delta \theta_{k,t}:=\theta_{k,t}-\theta_{k,t-1},\qquad
\Delta g_{k,t}:=g_{k,t}-g_{k,t-1}.
\end{equation}
To ensure the stability, we define a validity mask and the per-coordinate quotients,
\begin{equation}
\label{eq:h_update}
m_{k,t} :=
\begin{cases}
    1 & \text{if } |\Delta \theta_{k,t}| > \varepsilon \\
    0 & \text{if } |\Delta \theta_{k,t}| \le \varepsilon
\end{cases}
\qquad \qquad
h_{k,t}\;:=\;\frac{\Delta g_{k,t}}{\Delta \theta_{k,t}}\;\;\text{(element-wise)}.
\end{equation}
We use the following projection operator (or clamp function) to limit the range of computed quotients:
\begin{equation}
    \Pi_{[a,b]}(x)
    = \operatorname*{arg\,min}_{y \in [a,b]} |y - x|
    = \min\!\bigl(\max(x,a),\, b\bigr).
\end{equation}
The clamped diagonal entries of Hessian are computed as:
\begin{equation}
\label{eqn:hessian}
\widehat H_k\;:=\;
\Pi_{[\lambda_{\mathrm{min}},\,\lambda_{\mathrm{max}}]}
\!\left(
\frac{\sum_{t=1}^{T-1}t\cdot(m_{k,t}\odot h_{k,t})}
     {(\sum_{t=1}^{T-1}t\cdot m_{k,t})+\varepsilon}
\right)\!\in\mathbb{R}^d\text{ (element-wise division)}.
\end{equation}
where $0<\lambda_{\mathrm{min}}\le \lambda_{\mathrm{max}}<\infty$ and the division is
element-wise. The $t$-weights prioritize later inner steps whose finite-difference quotients are closer to the current state. At the same time, the averaging reduces mini-batch noise \rev{(that typically corrupts curvature information)} to yield a lower-variance (approximately unbiased) diagonal curvature estimate. This way, the diagonal entries provide direct information about local
curvature. In \opt\ they are applied \rev{at the end of each epoch as a correction step}\rem{as preconditioner \emph{at the end of each epoch}}
\begin{equation}
P_k \;=\; \mathbf{1} \oslash \widehat H_k\;=\; \big(1/H_{k,1},\ldots,1/H_{k,d}\big)\!\in\mathbb{R}^d.
\end{equation}
\emph{Note on Memory and Compute:} The approximations in Eqn.~(\ref{eqn:hessian}) can be calculated with rolling accumulators
for the (weighted) numerator and denominator, as well as the most recent $(\theta_{k,t},g_{k,t})$
to form $(\Delta \theta_{k,t},\Delta g_{k,t})$. Their computation only requires $O(d)$ memory per tensor and avoids history buffers.

\textbf{One Step Update each Epoch}
At the end of each epoch, we have the most accurate estimate of diagonal of Hessian, which provide us the direct information on local curvature. \opt\ performs one update:
\begin{equation}
\theta_{k+1,0}\;=\;\theta_{k,T-1}-\eta_2\,P_k\,\tilde g_k,
\end{equation}
where the gradients $\tilde g_k$ are computed as the rolling accumulation by reusing the same rolling accumulators (weighted sum of $g_{k,t}$ and sum of $t$), keeping memory overhead minimal:
\begin{equation}
\tilde g_k = 
\dfrac{\sum_{t=0}^{T-1}t\cdot g_{k,t}}{\sum_{t=0}^{T-1}t+\varepsilon}
\end{equation}
Alternatively, we can use the gradients from the last mini-batch:
\begin{equation}
\tilde g_k = 
g_{k,T-1},
\end{equation}

The first method reuses the same rolling accumulators (weighted sum of $g_{k,t}$ and sum of $t$), keeping memory overhead minimal; while using the gradients in the last mini-batch requires less computing but higher-variance. 

\textbf{Scaling Coefficient for the Next Epoch}
We calculate the scaling factor for the next epoch as:
\begin{equation}
\gamma_k\;:=\;
\!Q_{\omega}(\widehat H_k),
\end{equation}
Here $Q_{\omega}(\cdot)$ is \rem{taken}\rev{computed} over the entries of the per-tensor diagonal Hessian estimate $\widehat H_k$; i.e. we compute the low-tail $\omega$-quantile across its coordinates (elements). Using a low-tail quantile biases $\gamma_k$ toward small curvature directions (small diagonal entries), accelerates the effective learning rate when curvature is low, while remaining robust to outliers. 
In practice, $\gamma_k$ is computed separately for each tensor/layer, using that tensor’s diagonal Hessian estimate\rev{, and since $\widehat H_k$ is bounded by $\lambda_{mim}$ and $\lambda_{max}$ so is $\gamma_k$}; if the quantile is numerically undefined, we set $\gamma_k = 1$.

\textbf{Overall Algorithm} The overall algorithm of \opt\ is summarized below:
\begin{algorithm}[htbp]
\caption{\textsc{\opt}}
\begin{algorithmic}[1]
\Require $\theta_0\!\in\!\mathbb{R}^d$, epochs $K$, inner steps $T$, batches $B_t$, steps $\eta_1,\eta_2$, clamp $[\lambda_{\mathrm{min}},\lambda_{\mathrm{max}}]$, percentile $\omega$, $\varepsilon\!>\!0$, mode $\in\{\texttt{avg},\texttt{last}\}$, First-order (FO) backbone method.
\State $\gamma\gets 1$
\For{$k=0..K-1$}
  \State $S_{\mathrm{num}},S_{\mathrm{den}}\gets 0$;\; $\theta\gets \theta_0$
  \For{$t=0..T-1$}
    \State sample $B_t$;\; $g_t\gets \nabla f_B(\theta)$
    \State $\mu_t\gets \eta_1/\big(\gamma-\bigl(\gamma-1\bigr)\frac{t}{T}\bigl)$\quad $\theta\gets \mathrm{FO_{step}}(\theta,g_t,\mu_t)$
    \If{$t\ge 1$}
      \State $m\gets \mathbf{1}_{\{|\theta-\theta_{-}|\;>\;\varepsilon\}}$;\; $h_t\gets m\odot\frac{g_t-g_{-}}{\theta-\theta_{-}}$
      \State $S_{\mathrm{num}}\!\mathrel{+}= t\,h_t$;\; $S_{\mathrm{den}}\!\mathrel{+}= t\,m$
    \EndIf
    \State $\theta_{-}\gets \theta$;\; $g_{-}\gets g_t$
  \EndFor
  \State $\widehat H\gets \operatorname{\Pi}_{[\lambda_{\mathrm{min}},\lambda_{\mathrm{max}}]}\!\Big(\frac{S_{\mathrm{num}}}{S_{\mathrm{den}}+\varepsilon}\Big)$;\; $P\gets \mathbf{1} \oslash \widehat H$
  \State $\gamma\gets Q_\omega(\widehat H)$ (fallback $1$)
  \State $\tilde g\gets \big(\sum_{t=0}^{T-1} t\,g_t\big)\big/(\sum_{t=0}^{T-1} t+\varepsilon)$ if \texttt{avg} else $g_{T-1}$
  \State $\theta \gets \theta-\eta_2\,P\,\tilde g$
\EndFor
\end{algorithmic}
\end{algorithm}

\textbf{Updating Scheme when ADAM is used}
When the first-order method is ADAM, we replace the first-order step (\Eqref{eqn:dk_schedule}) with the bias-corrected ADAM update (\Eqref{eq:adam_update}) while leaving the curvature tracking and quantile scaling unchanged. 
The diagonal curvature estimate $\widehat H_k$ (and masking/clamping) is computed exactly as in the \textsc{SGD} case using $\{g_t\}$, and the per-tensor scaling coefficient remains $\gamma_k:=Q_\omega(\widehat H_k)$ (fallback $1$). The epoch-level \rem{preconditioner}\rev{diagonal estimate} is still $P \;=\; \mathbf{1} \oslash \widehat H_k$, and the outer update uses the stored gradients (not ADAM moments) for $\tilde g$:
\[
\tilde g \;:=\; \frac{\sum_{t=0}^{T-1} t\,g_t}{\sum_{t=0}^{T-1} t+\varepsilon}\;\;\text{if \texttt{avg},\quad else}\;\; \tilde g:=g_{T-1},
\qquad
\theta \leftarrow \theta - \eta_2\,P\,\tilde g.
\]
In practice, all quantiles and \rem{preconditioners}\rev{hessian diagonal estimates} are computed per tensor/layer.

\textbf{Hyperparameters} A complete list of hyper-parameters and their default values can be find in Supplemental \Secref{subsec:hyperparameters}.

\subsection{Convergence of \opt\ }
\label{subsec:proof}

We cast the convergence proof of \opt\ as outlined in Eqn.~(\ref{eqn:func_def}). Note that we use $x$ as the variable name instead of $\theta$ to avoid confusion.

\begin{equation}
\label{eqn:func_def}
f(x)\;=\;\frac1n\sum_{i=1}^n f_i(x),\qquad x\in\R^d,
\end{equation}
where each $f_i$ is convex and has $L_i$-Lipschitz gradient (i.e., is $L_i$-smooth).
Let $L_{\max}\coloneqq \max_i L_i$.
    Assume $f$ attains a minimum and let $x^\star\in\argmin f$ with $f^\star\coloneqq f(x^\star)$.
At iteration $t$, SGD samples $i_t\in\{0,\dots,T\}$ uniformly at random and performs
\begin{equation}
x_{t+1}\;=\;x_t-\eta_1\,\nabla f_{i_t}(x_t),
\end{equation}
where $(\eta_1)_{t\ge 0}$ are step sizes.

We will also use the (at-optimum) gradient noise level
\begin{equation}
\sigma_f^{\star}\;\coloneqq\;\mathrm{Var}\!\left(\nabla f_i(x^\star)\right)
=\E\bigl[\norm{\nabla f_i(x^\star)}^2\bigr]-\norm{\E[\nabla f_i(x^\star)]}^2
=\E\bigl[\norm{\nabla f_i(x^\star)}^2\bigr],
\end{equation}
since $\nabla f(x^\star)=\frac1n\sum_i \nabla f_i(x^\star)=0$ by optimality (convex and smooth).

\opt\ introduces effective step sizes
$\mu_{k,t}=\eta_1/\gamma_{k,t}$.
At the end of each epoch ($k=1,\dots,K$) we form a diagonal \rem{preconditioner} $P_k \;=\; \mathbf{1} \oslash \widehat H_k$ from a clamped diagonal secant proxy
$\widehat H_k\in[\lambda_{min}, \lambda_{max}]^d$ (element-wise), and take one \rem{preconditioned} stochastic step with a \emph{fresh} sample:
\begin{equation}
x_{k+1,0}=x_{k,T}-\eta_2\,P_k\,\tilde g_k,
\qquad
\tilde g_k:=\nabla f_{i}(x_{k,T}).
\end{equation}

\subsection{Storage and Computational Complexity}
\label{subsec:ssgd-complexity}

We summarize the storage and dominant per-iteration computational cost below. Note $d$ represents the total number of trainable parameters.
We use $s=\Delta \theta$ and $y=\Delta g$ uniformly: $s$ is the parameter change and $y$ the corresponding gradient change.
For L\mbox{-}BFGS, let $r$ be the history size (number of stored $(s_i,y_i)$ pairs, typically $r\in[5,20]$).

\begin{table}[htbp]
\scriptsize
{\centering
\setlength{\tabcolsep}{5pt}
\caption{\footnotesize{Per-step total complexity (including f--b passes).} $d$ = \#\ parameters; $r$ = L-BFGS history size; $K_t$ = \#\ closure (line-search) evaluations at step $t$.
\label{tab:ssgd-complexity-total}}
\begin{adjustbox}{max width=\linewidth}
\begin{tabular}{@{}l c c c p{0.55\linewidth}@{}}
\toprule
Algorithm & Total storage & Elem.\ ops / step & f--b passes / step & Accounting / notes \\
\midrule
SGD
& $O(d)$
& $O(d)$
& $1$
& Parameters only; no optimizer state ($\approx d$). \\

SGD\,+\,momentum
& $O(2\,d)$
& $O(2\,d)$
& $1$
& Parameters $+$ one moment buffer ($\approx 2d$). \\

Adam
& $O(3\,d)$
& $O(3\,d)$
& $1$
& Parameters $+$ two moment buffers $m,v$ ($\approx 3d$). \\

\opt$^\dagger$
& $O(5\,d)$
& $O(5\,d)$
& $1$
& Parameters $+$ four $d$-vectors (prev.\ $\theta$, prev.\ $g$, rolling means $h,\bar{g}$) ($\approx 5d$).  \\

L\mbox{-}BFGS$^\ddagger$
& $O(r\,d)$
& $O(r\,d)$
& $K_t$
& Parameters $+$ history of $r$ pairs $(s_i,y_i)$; two-loop recursion $\bigl(\approx (2r\!+\!1)d\bigr)$.  \\
\bottomrule
\end{tabular}
\end{adjustbox}

}

$^\dagger$ Extra \emph{per-epoch} work: one low-tail quantile per tensor of size $d_\ell$; with linear-time selection this is $O\!\big(\sum_\ell d_\ell\big)=O(d)$; with sorting it is $O\!\big(\sum_\ell d_\ell \log d_\ell\big)\ll O(d\log d)$ since no global sort over $d$ is performed (both are outside per-step costs).\\
$^\ddagger$ With a (possibly inexact) line search evaluating $K_t$ closures at step $t$, total f--b passes per step are $K_t$; with a fixed step or no line search, $K_t=1$.
\end{table}

%% file: sections/03_tool.tex
\section{A Lightweight Visualization Testbed for Optimizers}
\label{sec:testbed}

We use a simple and self-contained testbed to illustrate the effectiveness of the \opt. The testbed
is a quadratic equation, added by a number of Gaussian kernels. To mimic the stochasticity and
non-stationarity, at each step, additional random Gaussian noises are added to the quadratic equation itself, as well
as the magnitude, mean, and standard deviation of the Gaussian kernel.
The testbed mimics three key components of NN training: (i) dynamics of stochastic
mini-batch, (ii) non-stationarity across steps, and (iii) the generalization gap between train and
test. The testbed, which will be publicly available after the paper is published, can be used to
test various gradient-based optimizers. Its details are in Supplemental \Secref{subsec:landscape}.


\begin{figure}[htbp]
    \centering
    \includesvg[width=0.75\linewidth]{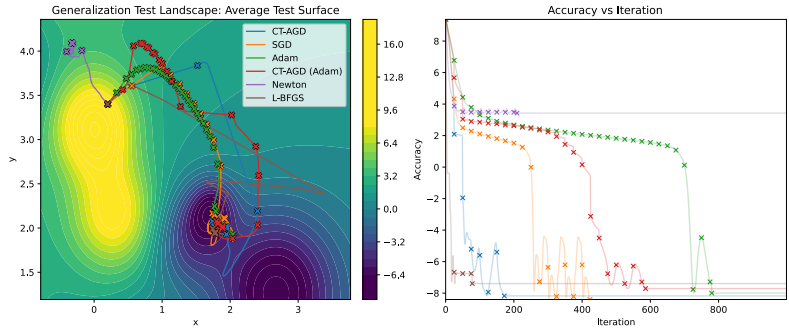}
    \caption{\footnotesize{LEFT: Illustration of convergence of \opt, SGD, Adam, Newton and L-BFGS where each
      step is shown. RIGHT: test accuracy versus iterations. The advange of \opt\ is clear.}}
    \label{fig:generalization_test_landscape_phase2}
\end{figure}

The trajectories in Fig.~\ref{fig:generalization_test_landscape_phase2} illustrate the convergence
of all methods in one instance. As shown, \opt, SGD, Adam, and L-BFGS descend into the common
basin where the minimum is located. The default \opt\ (with SGD as the backbone, first-order
method) requires the least number of steps to achieve so. \opt\ with ADAM as the backbone is also
faster than vanilla ADAM. The pure second-order method, Newton, failed to converge in this case due to its
poor tolerance to noise.

\begin{table}[ht]
\centering
\scriptsize
\setlength{\tabcolsep}{5pt}

\begin{tabular}{@{}l ccc ccc@{}}
\toprule
& \multicolumn{3}{c}{\textbf{With Random Components}} & \multicolumn{3}{c}{\textbf{Stationary}} \\
\cmidrule(lr){2-4}\cmidrule(lr){5-7}
\textbf{Optimizer} & Final value & \# Steps & Gen. gap & Final value & \# Steps & Time \\
\midrule
\opt\                    & $-4.10 \pm 1.29$ & $117.47 \pm 19.70$ & $2.31 \pm 0.91$ & $-6.33 \pm 1.61$ & $144.93 \pm 25.16$ & $0.49 \pm 0.08$ \\
\opt\,(Adam)            & $-4.61 \pm 1.13$ & $137.47 \pm 27.96$ & $1.66 \pm 0.90$ & $-5.75 \pm 2.15$ & $128.07 \pm 49.40$ & $0.43 \pm 0.16$ \\
SGD                     & $-3.93 \pm 1.10$ & $138.73 \pm 24.54$ & $2.81 \pm 0.55$ & $-6.52 \pm 1.56$ & $150.73 \pm 15.43$ & $0.52 \pm 0.06$ \\
Adam                    & $-4.31 \pm 1.30$ & $156.53 \pm 24.35$ & $2.08 \pm 0.76$ & $-5.49 \pm 2.33$ & $152.07 \pm 33.91$ & $0.54 \pm 0.12$ \\
L\mbox{-}BFGS & $-3.87 \pm 1.41$ & $278.93 \pm 189.92$& $2.15 \pm 0.97$ & $-5.19 \pm 2.48$ & $51.07 \pm 0.13$   & $0.33 \pm 0.14$ \\
\bottomrule
\end{tabular}
\caption{
  \label{tab::testbed}
  \footnotesize{Average of 15 testbed runs with different parameters for five
  method listed. Note that L\mbox{-}BFGS failed to converge in 3 runs, hence the average of 12 is
  listed. On the right, each run is stationary because no random Gaussian noises are added.}}
\end{table}

More comprehensive results are shown in 
Tab.~\ref{tab::testbed}. We repeated 15 testbed runs, each with a different set of parameters/initial points, and
recorded the optimization results as well as the number of steps required to converge. In addition, we evaluate the \emph{generalization gap}, which is the difference between average of test loss and average of training loss.

Observe that the first-order methods tend to achieve better answers (more
negative) than L-BFGS, and in fewer steps. Not surprisingly, L-BFGS requires
substantially more steps and appears less stable due to its poor tolerance to noise (failed to
converge in 3 runs). As a control, the results with no Gaussian random components (hence each run
becomes stationary) are listed on
the right. As expected, L-BFGS takes advantage of the Hessian information and converges much quicker.
However, its final value remains less accurate than the first-order methods. \opt\ with
different backbones converges in the fewest number of steps.

%% file: sections/04_results.tex
\section{Experimental Results}
\label{sec:results}

We compare \opt\ with SGD and Adam on three datasets: CIFAR-10, CIFAR-100 and
Tiny-ImageNet. For each case, we use three architectures: ResNet~\citep{He2016ResNet} and Wide
ResNet~\citep{Zagoruyko2016WRN} on all datasets. Additionally, a transformer architecture, DeiT~\citep{Touvron2021DeiT}, is used for
the Tiny-ImageNet dataset. Due to memory and computational constraints, we only run L-BFGS on the CIFAR-10
dataset using ResNet. Each run is repeated at least five times. For each experiment, we report (i) the best test accuracy and (ii) the number of epochs to
convergence, which is defined as the number of epochs to achieve within $5\%$ of the final
accuracy of that particular run. All methods share identical data pipelines and data
augmentation methods. All experiments are conducted with identical hardware and software
configurations, which are outlined in Supplemental \Secref{subsec:hw}. The hyperparameters used are listed in Supplemental \Secref{subsec:hyperparameters}. \rev{Finally, to demonstrate the method's effectiveness beyond computer vision, we provide additional evaluations on GraphSAGE in Supplemental \Secref{subsec:extra_experiments}.}

The results are summarized in Tab.~\ref{tab:cnn-all-compact-epochs}. The accuracy versus epoch of
three model-dataset pairs is also shown in Fig.~\ref{fig:curves-selected}. Our general observation
is that \opt\ achieves the same level of accuracy, but requires drastically fewer number of
epochs. A few outliers include TinyImageNet with ResNet-20. It appears that Adam is slightly
faster, however, the final testing accuracy is inferior ($44.28\%$ versus $46.24\%$ of \opt).

To provide a more reliable comparison, we present a longitudinal comparison in
Tab.~\ref{tab:convergence-only}. For each model-dataset pair, we set the cutoff test accuracy
as $5\%$ of the \emph{best} accuracy among all optimization methods. The table reports the number
of epochs for each optimization method to reach a given threshold. Note that some entries in Tab.~\ref{tab:convergence-only} are different from the corresponding values in Tab.~\ref{tab:cnn-all-compact-epochs} because of different accuracy thresholds.

\vspace{-.2cm}
\begin{table}[htbp]
    \centering
    \scriptsize
    \setlength{\tabcolsep}{4.5pt}
    \renewcommand{\arraystretch}{0.95}
    \begin{tabular}{@{}l l cc cc cc@{}}
        \toprule
        & & \multicolumn{2}{c}{\textbf{CIFAR-10}} & \multicolumn{2}{c}{\textbf{CIFAR-100}} & \multicolumn{2}{c}{\textbf{Tiny-ImageNet}} \\
        \cmidrule(lr){3-4}\cmidrule(lr){5-6}\cmidrule(lr){7-8}
        \textbf{Model} & \textbf{Optimizer} & Acc (\%) & \# Epochs & Acc (\%) & \# Epochs  & Acc
        (\%) & \# Epochs \\
        \midrule
        ResNet-20      & \opt\  & \cellcolor{curvesel} 90.05 $\pm$ 0.36 & \cellcolor{curvesel} \textbf{28.20 $\pm$ 2.22} & \textbf{64.70 $\pm$ 0.32} & \textbf{48.00 $\pm$ 2.91} & 46.24 $\pm$ 0.52 & 108.50 $\pm$ 7.50 \\
        ResNet-20      & SGD    & \cellcolor{curvesel} \textbf{90.35 $\pm$ 0.14} & \cellcolor{curvesel} 44.00 $\pm$ 3.29 & 64.10 $\pm$ 0.53 & 88.60 $\pm$ 7.43 & \textbf{46.88 $\pm$ 0.49} & 120.00 $\pm$ 8.91 \\
        ResNet-20      & Adam   & \cellcolor{curvesel} 89.20 $\pm$ 0.38 & \cellcolor{curvesel} 39.20 $\pm$ 3.09 & 61.86 $\pm$ 0.55 & 75.20 $\pm$ 2.69 & 44.28 $\pm$ 0.60 & \textbf{105.60 $\pm$ 7.78} \\

        ResNet-20      & \opt\ (Adam)&  88.82 $\pm$ 0.09 &  31.40 $\pm$ 2.08 & 61.57 $\pm$ 0.39 & 72.00 $\pm$ 6.74 & - & - \\
        ResNet-20      & L\,-\,BFGS&  82.79 $\pm$ 6.42 &  90.60 $\pm$ 29.50 & - & - & - & - \\
        \midrule
        WRN-28-4 & \opt\  & \textbf{93.68 $\pm$ 0.15} & \textbf{23.00 $\pm$ 0.88} & \cellcolor{curvesel} 73.84 $\pm$ 0.42 & \cellcolor{curvesel} \textbf{37.80 $\pm$ 3.66} & \textbf{58.69 $\pm$ 0.48} & \textbf{63.25 $\pm$ 5.42} \\
        WRN-28-4 & SGD    & \textbf{93.69 $\pm$ 0.19} & 33.80 $\pm$ 4.25 & \cellcolor{curvesel} \textbf{74.12 $\pm$ 0.44} & \cellcolor{curvesel} 67.80 $\pm$ 1.62 & \textbf{58.69 $\pm$ 0.50} & 96.40 $\pm$ 11.77 \\
        WRN-28-4 & Adam   & 93.35 $\pm$ 0.09 & 40.00 $\pm$ 7.02 & \cellcolor{curvesel} 74.05 $\pm$ 0.20 & \cellcolor{curvesel} 100.40 $\pm$ 2.26 & 57.11 $\pm$ 0.29 & 80.40 $\pm$ 5.99 \\

        WRN-28-4      & \opt\ (Adam)&  92.36 $\pm$ 0.23 &  23.00 $\pm$ 1.24 & 72.09 $\pm$ 0.24 & 45.20 $\pm$ 5.51 & - & - \\
        \midrule
        DeiT-12        & \opt\  & - & - & - & - & \cellcolor{curvesel} 31.87 $\pm$ 0.32 & \cellcolor{curvesel} \textbf{37.00 $\pm$ 2.32} \\
         DeiT-12        & SGD    & - & - & - & - & \cellcolor{curvesel} 31.31 $\pm$ 0.35 & \cellcolor{curvesel} 47.00 $\pm$ 8.14\\
        DeiT-12        & Adam   & - & - & - & - & \cellcolor{curvesel} \textbf{32.61 $\pm$ 0.31} & \cellcolor{curvesel} 115.40 $\pm$ 14.76 \\
        \bottomrule
    \end{tabular}
        \caption{
      \footnotesize{Test accuracy (\%) and number of epochs to convergence on three datasets with three model architectures. Best results for a model-data combination are in
      \textbf{bold}. Fig.~\ref{fig:curves-selected} shows the test accuracy versus epochs of
      shaded cells.}
    \label{tab:cnn-all-compact-epochs}}
\end{table}

\begin{figure*}[htbp]
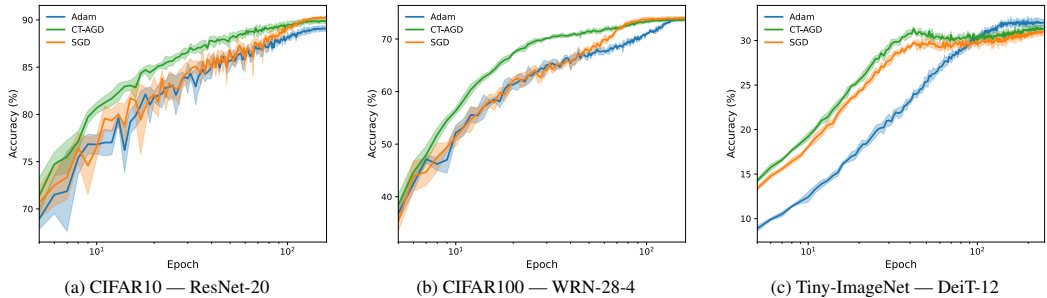

    \centering
    \setlength{\tabcolsep}{2pt}

    \begin{subfigure}[t]{0.32\linewidth}
    \centering
    \includesvg[width=\linewidth]{accuracy_cifar10_resnet}
    \caption{CIFAR10 — ResNet-20}
\end{subfigure}\hfill
    \begin{subfigure}[t]{0.32\linewidth}
    \centering
    \includesvg[width=\linewidth]{accuracy_cifar100_wideresnet}
    \caption{CIFAR100 — WRN-28-4}
\end{subfigure}\hfill
    \begin{subfigure}[t]{0.32\linewidth}
    \centering
    \includesvg[width=\linewidth]{accuracy_test_tiny_imagenet_deit}
    \caption{Tiny-ImageNet — DeiT-12}
\end{subfigure}

    \vspace{-0.25em}
    \caption{\footnotesize{Test accuracy versus epochs of selected model-dataset pairs. See
      Tab.~\ref{tab:cnn-all-compact-epochs} for more details.}
    \label{fig:curves-selected}}
\end{figure*}

\begin{table}[htbp]
    \centering
    \scriptsize
    \setlength{\tabcolsep}{5pt}
    \renewcommand{\arraystretch}{1.05}
    \begin{tabular}{@{}l l c c c c c c c@{}}
        \toprule
        \textbf{Model} & \textbf{Dataset} & \textbf{Acc. (\%)} 
        & \textbf{\opt} & \textbf{SGD} & \textbf{Adam} & \textbf{\opt\,(Adam)} & \textbf{Imp. 1}& \textbf{Imp. 2}\\
        \midrule
        ResNet-20  & CIFAR-10        & 85.7 & \textbf{24.0 $\pm$ 2.9} & 32.4 $\pm$ 3.4 & 39.8 $\pm$ 3.7 & 34.2 $\pm$ 3.4 & 25.9\% & 14.1\% \\
        ResNet-20  & CIFAR-100       & 61.2 & \textbf{42.2 $\pm$ 1.4} & 85.0 $\pm$ 5.0 & 117.8 $\pm$ 31.2 & 121.4 $\pm$ 16.8 & 50.4\% & -3.06\% \\
        ResNet-20  & Tiny-ImageNet   & 44.3 & 116.5 $\pm$ 5.5 & \textbf{113.0 $\pm$ 6.4} & 145.2 $\pm$ 17.7 & -- & -3.09\% & N/A\\
        \midrule
        WRN-28-4 & CIFAR-10       & 88.9 & \textbf{18.8 $\pm$ 1.6} & 22.8 $\pm$ 5.2 & 28.8 $\pm$ 5.2 & 23.6 $\pm$ 1.4 & 17.5\% & 18.1\% \\
        WRN-28-4 & CIFAR-100      & 70.2 & \textbf{31.2 $\pm$ 6.0} & 62.0 $\pm$ 3.6 & 83.8 $\pm$ 9.8 & 57.8 $\pm$ 3.2 & 49.7\% & 31.0\% \\
        WRN-28-4 & Tiny-ImageNet  & 55.5 & \textbf{54.5 $\pm$ 9.6} & 80.4 $\pm$ 13.8 & 87.8 $\pm$ 9.7 & -- & 32.2\% & N/A\\
        \midrule
        DeiT-12        & Tiny-ImageNet  & 30.4 & \textbf{35.6 $\pm$ 3.7} & 91.2 $\pm$ 49.6 & 96.2 $\pm$ 8.4 & -- & 61.0\% & N/A \\
        \midrule
        Average & & & & & & & \textbf{33.4\%} & \textbf{15.0\%}\\
        \bottomrule
    \end{tabular}
       \caption{\footnotesize{Number of epochs for a particular optimization method to reach a common cutoff
      threshold, which is shown in the third column. The best are shown in \textbf{bold}. The first improvement column represents the number of epochs improvement from SGD to \opt\ with SGD as the backbone. The second improvement column compares \opt\ with Adam as the backbone with Adam.
    \label{tab:convergence-only}}}
\end{table}
Overall we observe $33\%$ cases of fewer number of epochs to achieve the same accuracy. Note that we used the same hyperparameters for all experiments. In two cases when ResNet-20 is used, \opt\ is slightly slower than the baseline methods. However the difference is within $5\%$. Fine-tuning of hyperparameters can be a solution to move forward.  Also note the overfitting in DeiT, which is primarily driven by architectural/dataset. The issue is orthogonal to the optimizer, and should be addressed by other techniques such as data augmentation and regularization. Notice that \opt\ delivered significant reduction in terms of number of epochs. As shown in Section.~\ref{subsec:ssgd-complexity}, \opt\ requires slightly more computation per epoch. Our experiments show that per epoch runtime of \opt\ is about $9\%$ over its backbone method. However the overhead can be further reduced after code optimization. Overall \opt\ still provides significant net saving while achieving the same accuracy.

\rev{
\subsection{Ablation Studies}
\label{subsec:ablation}

We conduct ablation studies on CIFAR-10 (ResNet-20) using the default setup (same as in Tab.~\ref{tab:cnn-all-compact-epochs}). We focus on the main heuristic components of the
algorithm: (i) clamping of the diagonal curvature proxy, (ii) robustness to noisy curvature
quotients, (iii) curvature averaging and decay, and (iv) the low-tail quantile $\omega$.

\paragraph{Effect of clamping interval.}
Tab.~\ref{tab:ablation-interval-noise} reports a sweep of
$(\lambda_{\min},\lambda_{\max})$ spanning three orders of magnitude around the default
$(10^{-2},10^{2})$. 

\paragraph{Effect of computational noise on the curvature estimate.}
Tab.~\ref{tab:ablation-interval-noise} also reports experiments where we inject unbiased Gaussian noise at every mini-batch step. Here what we do is replace Eqn.~\ref{eq:h_update} with
\begin{equation}
    h_{k,t}\;:=\;\frac{\Delta g_{k,t}}{\Delta \theta_{k,t}} + \mathcal{N}(0,\sigma^2)
\end{equation}
where $\sigma^2$ is varied between experiments.

\begin{table}[H]
    \centering
    \scriptsize
    \setlength{\tabcolsep}{6pt}
    \renewcommand{\arraystretch}{.95}
    \begin{tabular}{@{}l c c c c c@{}}
        \toprule
        \textbf{Setting} & $\lambda_{\min}$ & $\lambda_{\max}$ & Noise $\sigma^2$ &
        \textbf{Top-1 (\%)} & \textbf{Epochs to conv.} \\
        \midrule
        Interval sweep & $10^{-1}$ & $10^{1}$ & 0 &
        $90.13 \pm 0.15$ & $24.60 \pm 2.86$ \\
        Interval sweep & $10^{-2}$ & $10^{2}$ & 0 &
        $90.01 \pm 0.22$ & $26.20 \pm 4.16$ \\
        Interval sweep & $10^{-3}$ & $10^{3}$ & 0 &
        $90.23 \pm 0.22$ & $25.80 \pm 1.25$ \\
        \midrule
        Noise sweep & -- & -- & $0$ &
        $90.09 \pm 0.16$ & $23.67 \pm 1.49$ \\
        Noise sweep & -- & -- & $0.01$ &
        $90.07 \pm 0.23$ & $25.80 \pm 2.39$ \\
        Noise sweep & -- & -- & $0.1$ &
        $90.12 \pm 0.14$ & $24.80 \pm 1.04$ \\
        \midrule
        SGD baseline & -- & -- & 0 &
        $90.35 \pm 0.14$ & $44.00 \pm 3.29$ \\
        \bottomrule
    \end{tabular}
    \caption{\footnotesize Ablations on the clamping interval
      $(\lambda_{\min},\lambda_{\max})$ and on injected curvature noise for \opt\ on
      CIFAR-10 with ResNet-20. The SGD row is reported for reference.}
    \label{tab:ablation-interval-noise}
\end{table}
\vspace{-.5cm}

The results show that both final accuracy and epochs to convergence remain essentially
unchanged across a broad range of $(\lambda_{\min},\lambda_{\max})$, and are also stable
under injected noise with variance up to $0.1$ at the level of individual
curvature quotients, which suggests that \opt\ is robust.
\vspace{-.2cm}
\paragraph{Annealing schedule, curvature averaging, and low-tail quantile.}
We next ablate the annealing schedule used for propagating curvature information across
iterations, the weighting scheme used to average curvature estimates within each epoch, and
the low-tail quantile $\omega$ that defines the effective scaling. The first block of
Tab.~\ref{tab:ablation-decay-quantile} compares: (i) removing annealing entirely by keeping $\mu_{k,t} = \eta_1 / \gamma_{k,t}$ constant through each epoch, (ii) an exponential annealing of the curvature cue with factor $\alpha = \tfrac{1}{2}$, (iii) the baseline linear annealing with weighted curvature estimation, and (iv) replacing the weighted
curvature average by a non-weighted average.

\begin{table}[htbp]
    \centering
    \scriptsize
    \setlength{\tabcolsep}{4.5pt}
    \renewcommand{\arraystretch}{.95}
    \begin{tabular}{@{}l c c@{}}
        \toprule
        \textbf{Setting} &
        \textbf{Top-1 (\%)} & \textbf{Epochs to conv.} \\
        \midrule
        No annealing &
        $50.96 \pm 3.82$ & $2.20 \pm 0.56$ \\
        Exponential annealing ($\alpha = 0.5$) &
        $90.32 \pm 0.35$ & $27.40 \pm 5.31$ \\
        Baseline &
        $90.01 \pm 0.22$ & $26.20 \pm 4.16$ \\
        Non-weighted curvature &
        $90.05 \pm 0.29$ & $26.80 \pm 1.62$ \\
        \midrule
        $\omega = 0.1$ &
        $90.23 \pm 0.22$ & $25.80 \pm 1.25$ \\
        $\omega = 0.2$ &
        $90.16 \pm 0.26$ & $23.80 \pm 3.21$ \\
        $\omega = 0.5$ &
        $90.18 \pm 0.11$ & $35.80 \pm 3.87$ \\
        \midrule
        SGD baseline &
        $90.35 \pm 0.14$ & $44.00 \pm 3.29$ \\
        \bottomrule
    \end{tabular}
    \caption{\footnotesize Ablations of annealing schedule, curvature averaging, and $\omega$ on CIFAR-10 with ResNet-20. }
    \label{tab:ablation-decay-quantile}
\end{table}
\vspace{-.3cm}

Removing annealing leads to severe degradation in accuracy and much higher
variance, showing that some form of annealing of the inter-epoch curvature cue is essential. This is consistent with our intuition that outdated curvature cues can hinder optimization steps.
Both exponential and linear annealing recover high accuracy and similar convergence speed,
while the non-weighted curvature average performs almost identically to the weighted
version. This indicates that \opt\ is sensitive to the presence of annealing, but not to a
precise functional form or a carefully engineered scheme.\\
The second block of Tab.~\ref{tab:ablation-decay-quantile} sweeps the low-tail quantile
$\omega$ used to define $\gamma_k$. Across a wide range of values ($0.1 \leq \omega \leq
0.5$), the final accuracy remains essentially unchanged, whereas the number of epochs to
convergence mainly reflects how aggressively the method exploits the curvature cue. Smaller
values of $\omega$ favor more aggressive updates and faster convergence.

}

%% file: sections/05_conclusion.tex
\section{Discussion}
\vspace{-.2cm}
The key insight of \opt\ is that with careful range limiting and other heuristics, direct calculation of the diagonals of the Hessian can 
benefit large-scale non-convex optimization problems, which frequently occur in deep neural network training. When compared to adaptive gradient methods (e.g., ADAM), which rely on the moments of the gradients to capture the local curvature, the direct approach deployed in \opt\ is more responsive to the change of the local curvature, thereby enabling convergence in a smaller number of epochs. Moreover, \opt\ can be interpreted as a \rev{coordinate-wise curvature informed learning rate modulator} \rem{curvature-aware preconditioning} approach, hence it can be deployed with other first-order optimization methods as its backbone. Our extensive experimental results show that on average \opt\ achieves in average $33\%$ fewer epochs for the same training task, with up to $61\%$ savings in the best case.\\
In certain extreme cases (especially with small models on challenging datasets), \opt\ is slightly slower than the baseline method. ResNet-20 on Tiny-ImageNet is one such example. Nevertheless, \opt\ still achieves comparable accuracy in this setting, in just $3\%$ more epochs. This little overhead may be reduced with modest hyper-parameter tuning; in practice, adjusting $\lambda_{\min}$, $\lambda_{\max}$, and the quantile threshold $\omega$ can be explored. \rev{That said, our ablation studies confirm the method’s robustness to these hyperparameter choices, minimizing the need for extensive tuning.}

%% file: sections/A1_appendix.tex
\section{Supplemental Material}

\subsection{Dynamic ``Generalization'' Landscape}
\label{subsec:landscape}

\textbf{Snapshots and drift.}
The landscape is defined by a sequence of \emph{train} snapshots and a shorter sequence of \emph{test} snapshots, built sequentially to emulate nonstationarity. Each snapshot is a sum of a convex quadratic baseline and a set of drifting Gaussian ``lumps'' with random signs (to create both attractive and repulsive features). Let $\theta\in\mathbb{R}^2$, let $c_0\in\mathbb{R}^2$ be the target center, $q>0$ the quadratic factor, and for $j=1,\dots,M$ let $(c_j,a_j,s_j,\sigma_j)$ denote center, amplitude, scale, and sign ($\sigma_j\in\{-1,+1\}$). The value of a snapshot $\mathcal{S}$ at $\theta$ is
\begin{equation}
    V(\theta;\mathcal{S})
    \;=\;
    q\,\|\theta-c_0\|_2^2
    \;+\;
    \sum_{j=1}^M \sigma_j\, a_j \exp\!\Bigl(-\tfrac{\|\theta-c_j\|_2^2}{2 s_j^2}\Bigr).
    \label{eq:landscape}
\end{equation}
Snapshots evolve by small stochastic drifts:
\begin{align*}
    c_j^{(t+1)} &= c_j^{(t)} + \varepsilon^{(t)}_{c,j}, 
    & \varepsilon^{(t)}_{c,j} &\sim \mathcal{N}(0,\sigma_c^2 I_2), \\
    a_j^{(t+1)} &= a_j^{(t)}\,[1+\varepsilon^{(t)}_{a,j}],
    & \varepsilon^{(t)}_{a,j} &\sim \mathcal{N}(0,\sigma_a^2), \\
    s_j^{(t+1)} &= s_j^{(t)}\,[1+\varepsilon^{(t)}_{s,j}],
    & \varepsilon^{(t)}_{s,j} &\sim \mathcal{N}(0,\sigma_s^2).
\end{align*}
Calling the landscape once advances the train snapshot index (cyclically), so an optimiser observes a \emph{changing} objective during a trajectory, akin to iterating over mini-batches and data augmentations.

\textbf{Train, test, and the gap.}
Let $\{\mathcal{S}^{\mathrm{train}}_t\}_{t=1}^{T_{\mathrm{train}}}$ and $\{\mathcal{S}^{\mathrm{test}}_u\}_{u=1}^{T_{\mathrm{test}}}$ be the built sequences. For any point $\theta$, we define
\begin{equation}
\begin{gathered}
\begin{alignedat}{2}
L_{\mathrm{avg}}(\theta) &:= \frac{1}{T_{\mathrm{train}}}\sum_{t=1}^{T_{\mathrm{train}}} V\!\left(\theta;\mathcal{S}^{\mathrm{train}}_t\right) \quad &
E_{\mathrm{avg}}(\theta) &:= \frac{1}{T_{\mathrm{test}}}\sum_{u=1}^{T_{\mathrm{test}}} V\!\left(\theta;\mathcal{S}^{\mathrm{test}}_u\right)
\end{alignedat} \\
\mathrm{Gap}(\theta) := E_{\mathrm{avg}}(\theta)-L_{\mathrm{avg}}(\theta)
\end{gathered}
\end{equation}

which play the role of train loss, test loss, and generalisation gap. In visualisations, we may draw either the instantaneous train surface $V(\cdot;\mathcal{S}^{\mathrm{train}}_t)$ or the averaged test surface $E_{\mathrm{avg}}(\cdot)$ as the background to contrast optimisation progress and generalisation.

To visualise the animation of how the landscape works and how different optimisers behave (results from Figure \ref{fig:generalization_test_landscape_phase2}), follow the link:\\ \href{https://osf.io/63zem?view_only=b0f43a3664f44dba98cad7a1c4c33cc2}{\textit{https://osf.io/63zem?view\_only$=$b0f43a3664f44dba98cad7a1c4c33cc2}}

\subsection{Evolution of the curvature-aware divisor}
\label{subsec:curvature_gamma}
\begin{figure}[htbp]
    \centering
    \includesvg[width=0.5\linewidth]{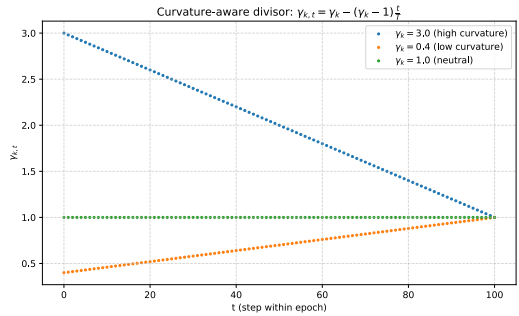}
    \caption{Evolution of the curvature-aware divisor \(\gamma_{k,t}=\gamma_k-(\gamma_k-1)\tfrac{t}{T}\) over one epoch (of 100 internal steps) for three initial values: \(\gamma_k>1\), \(0<\gamma_k<1\), and \(\gamma_k=1\). The schedule is linear from \(\gamma_k\) at \(t=0\) to \(1\) at \(t=T\).}
    \label{fig:gamma_evolution}
\end{figure}

\subsection{Aditional results}
\label{subsec:additional_results}

\begin{table}[htbp]
    \centering
    \scriptsize
    \setlength{\tabcolsep}{4.5pt}
    \renewcommand{\arraystretch}{0.95}
    \label{tab:cnn-all-compact-time}
    \begin{tabular}{@{}l l cc cc cc@{}}
        \toprule
        & & \multicolumn{2}{c}{\textbf{CIFAR-10}} & \multicolumn{2}{c}{\textbf{CIFAR-100}} & \multicolumn{2}{c}{\textbf{Tiny-ImageNet}} \\
        \cmidrule(lr){3-4}\cmidrule(lr){5-6}\cmidrule(lr){7-8}
        \textbf{Model} & \textbf{Optimizer} & Acc (\%) & Time (s) & Acc (\%) & Time (s) & Acc (\%) & Time (s) \\
        \midrule
        ResNet-20      & \opt\     &  90.05 $\pm$ 0.36 &  \textbf{255.13 $\pm$ 30.17} & \cellcolor{curvesel} \textbf{64.70 $\pm$ 0.32} & \cellcolor{curvesel} \textbf{463.03 $\pm$ 15.86} & \cellcolor{curvesel} 46.24 $\pm$ 0.52 & \cellcolor{curvesel} 4343.79 $\pm$ 207.09 \\
        ResNet-20      & SGD       &  \textbf{90.35 $\pm$ 0.14} &  316.81 $\pm$ 28.85 & \cellcolor{curvesel} 64.10 $\pm$ 0.53 & \cellcolor{curvesel} 857.62 $\pm$ 54.83 & \cellcolor{curvesel} \textbf{46.88 $\pm$ 0.49} & \cellcolor{curvesel} \textbf{3660.27 $\pm$ 202.14} \\
        ResNet-20      & Adam      & 89.20 $\pm$ 0.38 & 388.12 $\pm$ 37.54 & \cellcolor{curvesel} 61.86 $\pm$ 0.55 & \cellcolor{curvesel} 1179.86 $\pm$ 306.30 & \cellcolor{curvesel}  44.28 $\pm$ 0.60 & \cellcolor{curvesel} 4771.47 $\pm$ 660.88 \\
        ResNet-20 & \opt\ (Adam) &  88.82 $\pm$ 0.09 &  374.33 $\pm$ 36.40 & 61.57 $\pm$ 0.39 & 1335.25 $\pm$ 190.49 &  - & - \\
        ResNet-20      & L\,-\,BFGS& 82.79 $\pm$ 6.42 & 16295.83 $\pm$ 3140.84 &  - &  - &  - &  - \\
        \midrule
        Wide ResNet    & \opt\     & \cellcolor{curvesel} 93.68 $\pm$ 0.15 & \cellcolor{curvesel} \textbf{388.66 $\pm$ 33.75} &  73.84 $\pm$ 0.42 &  \textbf{666.94 $\pm$ 132.49} & \cellcolor{curvesel} \textbf{58.69 $\pm$ 0.48} & \cellcolor{curvesel} \textbf{9382.40 $\pm$ 2322.14} \\
        Wide ResNet    & SGD       & \cellcolor{curvesel} \textbf{93.69 $\pm$ 0.19} & \cellcolor{curvesel} 416.61 $\pm$ 95.57 &   \textbf{74.12 $\pm$ 0.44} &  1140.35 $\pm$ 66.81 & \cellcolor{curvesel} \textbf{58.69 $\pm$ 0.50} & \cellcolor{curvesel} 11015.84 $\pm$ 1881.44 \\
        Wide ResNet    & Adam      & \cellcolor{curvesel} 93.35 $\pm$ 0.09 & \cellcolor{curvesel} 525.29 $\pm$ 93.24 & 74.05 $\pm$ 0.20 &  1542.87 $\pm$ 185.55 & \cellcolor{curvesel} 57.11 $\pm$ 0.29 & \cellcolor{curvesel} 12000.42 $\pm$ 1317.28 \\
        Wide ResNet & \opt\ (Adam) &  92.36 $\pm$ 0.23 &  524.07 $\pm$ 31.57 &  72.09 $\pm$ 0.24 &  1283.09 $\pm$ 72.73 &  - &  - \\
        \midrule
        DeiT           & \opt\     & - & - & - & - &  31.87 $\pm$ 0.32 &  \textbf{3785.37 $\pm$ 415.78} \\
        DeiT           & SGD       & - & - & - & - &  31.31 $\pm$ 0.35 &  8598.44 $\pm$ 4677.69 \\
        DeiT           & Adam      & - & - & - & - &  \textbf{32.61 $\pm$ 0.31} &  9132.21 $\pm$ 799.30 \\
        \bottomrule
    \end{tabular}
    \caption{
         Test accuracy (\%) and \emph{Time to convergence}. Accuracies are the same as in Tab.~\ref{tab:cnn-all-compact-epochs}, but convergence is defined as the first epoch at which a run reaches 5\% of the \emph{best} max test accuracy among optimizers for the same model–dataset pair. Best results for a model-data combination are in \textbf{bold}. Fig.~\ref{fig:curves-remaining} shows the test accuracy versus epochs of shaded cells.}
\end{table}

\begin{figure}[htpb]
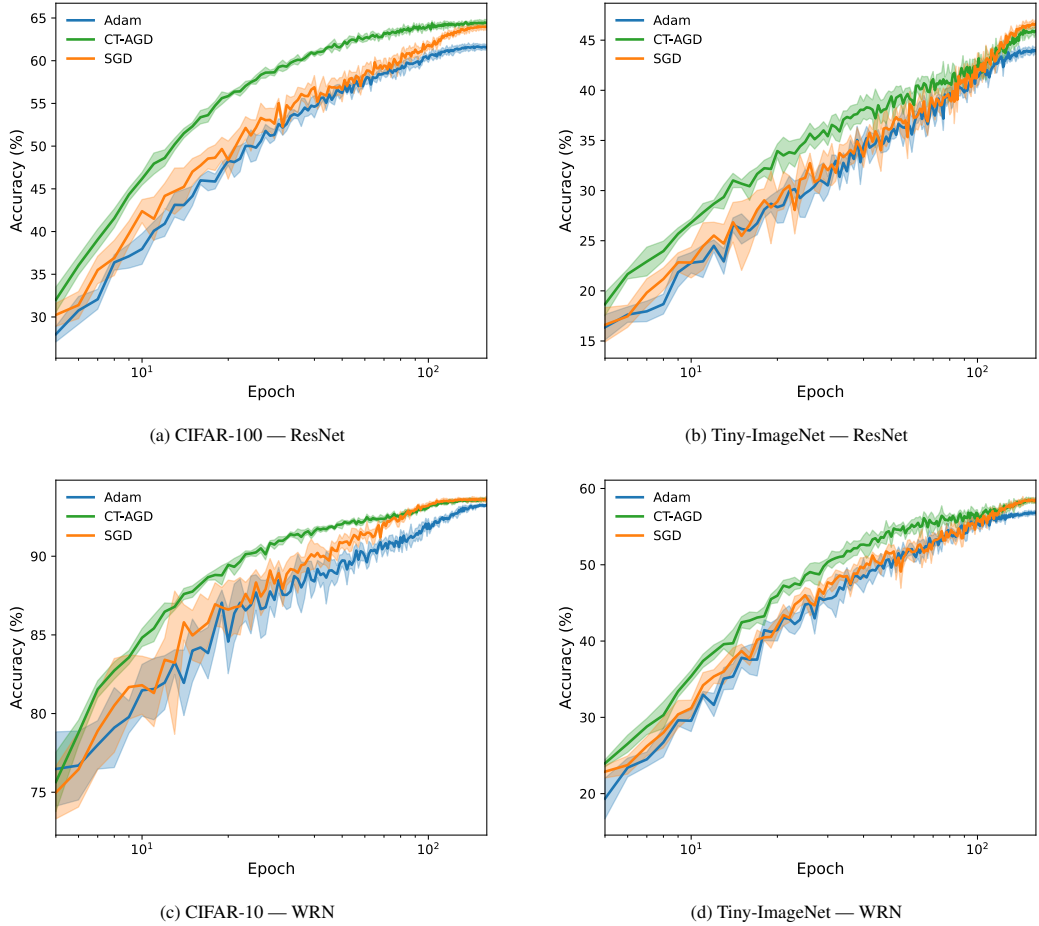

    \centering
    \captionsetup[subfigure]{font=scriptsize,skip=2pt}

    \begin{subfigure}[t]{0.48\linewidth}
        \centering
        \includesvg[width=\linewidth]{accuracy_cifar100_resnet}
        \caption{CIFAR-100 — ResNet}
    \end{subfigure}\hfill
    \begin{subfigure}[t]{0.48\linewidth}
        \centering
        \includesvg[width=\linewidth]{accuracy_tiny_imagenet_resnet}
        \caption{Tiny-ImageNet — ResNet}
    \end{subfigure}

    \vspace{0.6em}

    \begin{subfigure}[t]{0.48\linewidth}
        \centering
        \includesvg[width=\linewidth]{accuracy_cifar10_wideresnet}
        \caption{CIFAR-10 — WRN}
    \end{subfigure}\hfill
    \begin{subfigure}[t]{0.48\linewidth}
        \centering
        \includesvg[width=\linewidth]{accuracy_tiny_imagenet_wideresnet}
        \caption{Tiny-ImageNet — WRN}
    \end{subfigure}

    \vspace{-0.5em}
    \caption{\textbf{Additional accuracy trajectories.} Complements Fig.~\ref{fig:curves-selected} by showing the non-diagonal tasks.}
    \label{fig:curves-remaining}
\end{figure}

\rev{
\subsection{Additional Experiments}
\label{subsec:extra_experiments}

We expanded our evaluation by testing \opt\ on large-scale node classification on the Reddit benchmark.
The Reddit graph contains $\sim 232\mathrm{k}$ nodes and $\sim11.6\mathrm{M}$ edges with $602$-dimensional input features and $41$ target classes.
We train a 2-layer GraphSAGE over $60$ epochs.
All optimizers share the same architecture and training protocol.
Tab.~\ref{tab:extra-reddit} reports the final training accuracy and the number of epochs to convergence, averaged over $5$ runs.

    \begin{table}[h]
    \centering
    \scriptsize
    \setlength{\tabcolsep}{6pt}
    \renewcommand{\arraystretch}{1.05}
    \begin{tabular}{@{}lcc@{}}
    \toprule
    \textbf{Method} & \textbf{Train acc. (\%)} & \textbf{Epochs to conv.} \\
    \midrule
    Adam  & $97.60 \pm 0.00$ & $23.4 \pm 0.5$ \\
    CT-AGD (Adam) & $97.60 \pm 0.00$ & $23.2 \pm 0.4$ \\    
    CT-AGD (SGD) & $93.30 \pm 0.00$ & $18.8 \pm 0.8$ \\
    SGD  & $92.80 \pm 0.00$ & $27.8 \pm 0.4$ \\
    \bottomrule
    \end{tabular}
    \caption{\footnotesize Reddit node classification with a 2-layer GraphSAGE encoder.
    CT-AGD (Adam) matches the strong training accuracy of Adam with essentially identical convergence,
    while CT-AGD (SGD) improves over plain SGD in both training accuracy and the number of epochs required to converge.}
    \label{tab:extra-reddit}
    \end{table}

}

\subsection{Hardware and Software}
\label{subsec:hw}
Hardware and software configurations for the experimental results.
\begin{table}[H]
\centering
\renewcommand{\arraystretch}{1.12}
\begin{tabular}{@{}ll@{}}
\toprule
\multicolumn{2}{@{}l}{Hardware} \\
\midrule
CPU                & Intel Xeon Silver 4214R @ 2.40\,GHz \\
vCPU topology      & 8 vCPUs; 2 threads/core; NUMA nodes: 2 \\
Memory             & 64\,GiB RAM (swap: 0\,B); HugePages: 0 \\
GPU                & NVIDIA Tesla V100S-PCIe-32GB (1$\times$; 32\,GB HBM2) \\
Driver             & NVIDIA 555.42.02 \\
\addlinespace
\midrule
\multicolumn{2}{@{}l}{Software} \\
\midrule
OS / Kernel        & Ubuntu 20.04.4 LTS (Focal), Linux 5.4.0-120-generic \\
Python             & 3.10.18 \\
PyTorch            & 2.7.1\,+cu126 \\
Torchvision        & 0.22.1 \\
CUDA               & 12.6 \\
\bottomrule
\end{tabular}
\end{table}

\subsection{Hyper-parameters}
\label{subsec:hyperparameters}

\paragraph*{\normalfont \opt\ hyper-parameters and values used}
\begin{itemize}
  \item Secondary learning rate: \(\eta_2 = 0.5\).
  \item Clamp bounds: \(\lambda_{\mathrm{\min}} = 10^{-2}\), \(\lambda_{\mathrm{\max}} = 10^{2}\).
  \item Percentile statistic: \(\omega = 0.1\).
  \item Numerical stabilizer: \(\varepsilon = 10^{-3}\).
  \item Gradient update mode: \texttt{avg}.
\end{itemize}

\opt\ inherits the hyperparameters of its backbone optimizer:
\opt\ (Adam) uses Adam’s defaults; \opt\ uses SGD’s defaults. This means, they use the same values as the standalone optimizer.
\begin{center}
\begin{tabular}{l l}
\hline
Method & Primary learning rate \(\eta_1\) and common defaults \\
\hline
Adam and \opt\ (Adam) & \(\eta_1=10^{-3}\); \((\beta_1,\beta_2)=(0.9, 0.999)\) \\
SGD and \opt\  & \(\eta_1=10^{-2}\); momentum \(\mu=0.85\) (no Nesterov) \\
L-BFGS & initial step size \(\eta_1=1.0\); history size \(r=10\); max inner iterations \(K=20\) \\
\hline
\end{tabular}
\end{center}
All methods use weight decay \(\lambda = 5 \times 10^{-4}\).